%% file: template.tex
\documentclass[a4paper]{article}
\usepackage{enumitem}
\usepackage{multirow}
\usepackage{INTERSPEECH2019}
\usepackage{tabularx}
\usepackage{float}
\usepackage{comment}
\usepackage[title]{appendix}
\ninept

\title{Towards Universal Dialogue Act Tagging for Task-Oriented Dialogues} 
\name{Shachi Paul, Rahul Goel, Dilek Hakkani-T\"ur}

\address{Amazon}
\email{{\tt \{shachp, goerahul, hakkanit\}@amazon.com}}

\begin{document}
\maketitle
\vspace{-2mm}
\begin{abstract}
Machine learning approaches for building task-oriented dialogue systems require
large conversational datasets with labels to train on.  We are interested in
building task-oriented dialogue systems from human-human conversations, which
may be available in ample amounts in existing customer care center logs or can
be collected from crowd workers.  Annotating these datasets can be prohibitively
expensive.  Recently multiple annotated task-oriented human-machine dialogue datasets
 have been released, however their annotation schema varies across
different collections, even for well-defined categories such as dialogue acts
(DAs). We propose a Universal DA schema for task-oriented dialogues and align
existing annotated datasets with our schema. Our aim is to train a Universal DA
tagger (U-DAT) for task-oriented dialogues and use it for tagging human-human
conversations. We investigate multiple datasets, propose manual and automated
approaches for aligning the different schema, and present results on a target
corpus of human-human dialogues. In unsupervised learning experiments we
achieve an F1 score of 54.1\% on system turns in human-human dialogues. 
In a semi-supervised setup, the F1 score increases to 57.7\% which
would otherwise require at least 1.7K manually annotated turns. For new domains, we show
further improvements when unlabeled or labeled target domain data is
available. 
\end{abstract}
\vspace*{0pt}

\input{intro}

\input{related}

\input{approach}

\input{data}

\input{experiments}
\input{alignment}

\input{results}

\input{conclusion}

\bibliographystyle{IEEEtran}

\bibliography{mybib}
\input{appendix}

\end{document}

%% file: intro.tex
\section{Introduction}
\label{sec:intro}

\vspace*{0pt}
Dialogue acts (DAs) aim to portray the meaning of utterances at the
level of illocutionary force, capturing a speaker's intention in
producing that utterance~\cite{Austin:1975}. DAs have been
investigated by dialogue researchers for many
years~\cite{stolcke2000dialogue} and multiple taxonomies have been
proposed~\cite{anderson1991hcrc,core1997coding,bunt2009dit++} (see
~\cite{mezza2018iso} for a review).  Recent work in task-oriented
dialogue systems proposed a set of core DAs that describe interactions
at the level of intentions~\cite{young2007cued, Bunt2010towards, Shah:2018}.
With these, the system actions as output by a dialogue system policy
are commonly represented as the system DAs and associated
entities~\cite{gavsic2009back}.  Previous work on dialogue policy
learning and end-to-end training of dialogue systems rely on
supervised learning approaches to estimate system actions at each
turn, given the dialogue state or the previous conversation. These
models can then be fine-tuned with reinforcement
learning~\cite{Williams:2017, BingLiu:2018}.

In this work, we build an RNN-based DA tagger for tagging human-human
task-oriented conversations with DAs from a Universal DA schema that is
representative of the commonly-used acts in task-oriented dialogue systems. Our
long term goal is to use these annotated human-human dialogues to train
end-to-end dialogue systems to predict system actions for new dialogue-task
domains. Here, we focus on automatically annotating system-side DAs on
human-human dialogues. Such human-human dialogues for the new domain can be
found in existing customer care center logs or collected via crowdsourcing by
pairing two crowdworkers~\cite{multiwoz} or asking a single crowd worker to
write self dialogues~\cite{krause2017edina}. Previous work on DA tagging mainly
focused on human-human social interactions, such as the Switchboard
corpus~\cite{switchboard}, with little or no attention to task-oriented
dialogues.

Recently, multiple annotated task-oriented human-machine dialogue datasets have
been released~\cite{dstc-2,Shah:2018}, fostering research in this area. Hence,
we focus on learning to tag DAs from these human-machine dialogues, and applying
the learned models to human-human dialogues for task-oriented systems.  However,
the annotation schema varies across different corpora, even for well-defined
categories, such as DAs. Towards this goal we experiment with various alignment
schemes, propose a Universal schema of DAs across multiple existing corpora,
align the corpora accordingly to train a Universal DA tagger (U-DAT).

We use U-DAT to tag human-human multi-domain dialogues
(MultiWOZ-2.0~\cite{multiwoz}).  In our semi-supervised learning experiments we
achieve an F1 score of 57.7\% on system-turns on human-human data, which
requires at least 1.7K manually annotated turns.  We examine the potential of
domain adaptation of the U-DAT by leave-one-domain-out experiments. In presence
of a new domain we compare the performance of DA tagging using unsupervised
(w.r.t. target corpus), semi-supervised (self-training) and supervised
approaches. For these domains, we show further improvements when unlabeled or
labeled target domain data is available, providing guidelines on bootstrapping a
new domain without any DA annotations.

Our work has multiple novel contributions including a new hierarchical recurrent
neural network based approach for tagging DAs, a Universal DA schema for
task-oriented dialogues, alignment of multiple datasets to the universal schema,
using the aligned corpus for training of U-DAT for human-human dialogue
annotation and showcasing alternatives when bootstrapping a new domain.


\vspace*{0pt}

%% file: related.tex
\vspace{-2mm}
\section{Related Work}
\label{sec:related}
\vspace*{0pt}
The mismatch between multiple DA taxonomies has been identified
by~\cite{mezza2018iso} previously, where a subset of ISO 24617-2 (the
international ISO standard for DA annotation) tags~\cite{Bunt:2011}
have been identified and annotations of multiple corpora were mapped
to this set, focusing on social conversations. Our work has a similar
goal, but focuses on DAs that are necessary for task completion in
task-oriented interactions.

Since the publication of the seminal work on a machine learning approach for DA
tagging \cite{stolcke2000dialogue}, multiple learning approaches have been
proposed for this task, including maximum entropy taggers~\cite{ME-DA},
conditional random fields~\cite{CRF-DA}, and dynamic Bayesian
networks~\cite{DBN-DA}. Recent studies investigated recurrent and convolutional
neural networks with a pooling layer for short-text classification tasks, such
as DA tagging~\cite{lee2016sequential}. But these works don't take into account
the dialogue context. However, in a task-oriented conversation, there is a
strong correlation between system and user acts. For example, a user
usually \textit{inform}s when a system \textit{request}s information. Our work
represents short user utterances using recurrent neural networks, and
additionally models dialogue context using a hierarchical recurrent neural
network. Such dialogue-level models have also been proposed
in~\cite{liu2017using} for dialogue act tagging of human-human social phone
conversations. Previous studies mainly considered DA tagging of multi-human
conversations, such as the Switchboard~\cite{switchboard} corpus and meetings,
such as the ICSI meeting corpus~\cite{MRDA}, whereas our focus lies on modeling
system-side DAs. In dialogue systems, the system utterances are also generated
from system actions and are hence, observable. Thus, in our context
representation we include past system DAs in addition to system utterances. For
user utterances as well as system-side DAs in human-human conversations, we use
the predicted DAs.
Domain adaptation of DA tagging with unlabeled data was also investigated
by~\cite{margolis2010domain} for two human-human conversation genre, telephone
speech and face-to-face meetings. However, that work did not have annotation
mismatch issues across different datasets.
\vspace*{0pt}

%% file: approach.tex
\vspace*{-2ex}
\section{DA Tagging for Dialogue Systems}
\vspace*{0pt}
Let a dialogue $D$ with $N$ turns be denoted as a series of user and system
utterances, $u_i$, i.e. $D$ = $u_1, u_2, ...,u_{N}$ and $A$ be the predefined set of $M$ DAs i.e. $A = a_1,a_2...,a_M$. Given an utterance $u_i$
and its conversation history, DA tagging aims to predict the
set of DAs $A_i \subset A$ of $u_i$.
\vspace*{0pt}

\vspace*{0pt} We use a deep neural network based model for DA tagging. The
input to the model is the utterance $u_i$ and the conversation context
$C_i$ which is a function of the past utterances and their
corresponding DAs i.e. $C_i = f((u_1, A_1),...,(u_{i-1}, A_{i-1}))$. Since the utterance $u_i$ can be classified
into one or more DAs in $A$, the model makes a
binary-decision $y_j \in [0,1]$ for every DA ~$a_j \in
A_{1..M}$ for candidature.
\vspace*{0pt}

\vspace*{0pt}
For a dialogue $D_k$ and every DA class $a_j \in A$, we minimize the
following cross-entropy loss:
\vspace*{0pt}

\vspace{-2mm}
\begin{equation}
  L =  -\sum_{j=1}^{M}\sum_{i=1}^{N}\log\left(P\left(y_j|C_i^k,u_i^k\right)\right).
\end{equation}
\vspace{-2mm}

We use the following encoders to represent context:
\vspace{-2mm}
\begin{enumerate}
  \itemsep -1ex
  \vspace*{0pt}
	\item A bi-directional LSTM to encode each utterance,
          $u_i=w_1^i,...,w_{n_i}^i$, where $n_i$ denotes the number of tokens in
          $u_i$ and the final utterance representation for utterance $z_i$ is
          obtained by concatenating the last hidden layer of the forward LSTM,
          $\overrightarrow{LSTM}$, and the first hidden layer of the backward
          LSTM, $\overleftarrow{LSTM}$:
          \vspace{-2mm}
$$z_i = \overleftarrow{LSTM}(u_i) \oplus \overrightarrow{LSTM}(u_i)$$
\vspace{-4mm}
	\item A hierarchical, uni-directional LSTM to encode the dialogue level
          information, $e_i$:
          \vspace{-2mm}
$$ \vspace{-1mm} e_i = LSTM(z_1,...z_{i-1}) \vspace{-1mm}$$
\vspace{-4mm}
	\item An indicator number, $g_i$, representing whether the agent is user or the system, 
          i.e., $g_i=0$, if $u_i^{agent} = user$, $g_i=1$ otherwise. 
	\item Encoding over past DA(s) $p_i$, where the final
          representation is obtained by concatenating the many-hot
          representations of past-DAs. A DA vector is represented
          as a many-hot vector $d_i$ of dimension M, where we mark the true DAs as 1. 
          \vspace{-2mm}
	$$ p_i = d_1 \oplus d_2 \oplus ... \oplus d_{i-1} $$
 \vspace{-2mm}
\end{enumerate}
\vspace{-5mm}

The final encoded context $C_i$ is given by:
\vspace{-3mm}
\begin{equation}
\vspace{-1mm}
  C_i = e_i \oplus g_i \oplus p_i
\vspace{-1mm}
\end{equation}
\vspace*{0pt}
$C_i$ is then fed into a feed forward network $FF_j$, along with $u_i$, for each DA. The context encoders are shared for all acts.
\vspace*{0pt}
\vspace{-2mm}
\begin{equation}
  y_j = sigmoid(FF_j(z_i \oplus C_i))
\end{equation}
\vspace{-7mm}

%% file: data.tex
\vspace{0mm}
\section{Datasets and Experiments}\label{sec:data} 

\vspace*{0pt}
Our aim is to train a Universal DA tagger using public datasets, but the label
spaces across these datasets are not aligned. Therefore, we need a unified
representation of all the acts present across the datasets. We obtain this
representation by manually going through the datasets and aligning semantically
similar sentences to the same DA.
\vspace*{0pt} We chose the Google Simulated Dialogue (GSim) dataset
\cite{Shah:2018} and the DSTC2 dataset~\cite{dstc-2} for our
experimentation as they are both inspired by the CUED schema
\cite{young2007cued} for DAs. The GSim data has two parts
and was collected by generating dialogue flows for movie (GSim-M) and
restaurant (GSim-R) booking domains, where the individual turns from
simulation in terms of DAs and associated arguments were
then converted to natural language by crowd workers. DSTC2 contains
human-machine interactions collected for the second dialogue state
tracking challenge~\cite{dstc-2}. To experiment with DA
tagging on human-human conversational interactions, we use the
MultiWOZ-2.0~\cite{multiwoz}, which was collected by assigning tasks
(such as, booking a restaurant table and a cab to get there) and roles
(such as, user and agent) to two crowd workers, paired to accomplish
the task. The three datasets are summarized and compared across
various metrics in Table~\ref{tab:data_stats}. The last two rows of
the table show the vocabulary size of the system turns and unique
number and percentage of system turns after delexicalization, which
replaces the entity values with an entity type. The percentage of unique
turns is obtained by dividing the unique number of system turns with
the total number of system turns. Since the GSim and MultiWOZ-2.0
dialogues were written by crowdworkers, they include lots of
variation in the output system turns, whereas DSTC2 system turns were
generated by the participating systems, and have much less richness
for building DA taggers for system acts, but provides consistent annotations. 

\vspace*{0pt}
\begin{table}
\caption{Data statistics of various datasets}\label{tab:data_stats}
\vspace{-4mm}
\resizebox{\columnwidth}{!}{
  \begin{tabular}{l|r|r|r||r}
Data Sets: & GSim-R & GSim-M & DSTC2 & MultiWOZ-2.0\\
\hline
\# Dialogues Train & 1,116 & 384 & 1,612 & 8,438\\
\# Dialogues Dev & 349 & 120 & 506 & 1,000\\
\# Dialogues Test & 775 & 264 & 1,117 & 1,000\\
Avg \# Turns/Dialogue & 5.5 & 5.1 & 7.2 &6.7\\
\#  Sys Dialogue Acts & 7 & 7 & 12 & 14\\
SysTurn Vocab Size & 577 & 349 & 229& 15,408\\
\#Uniq SysTurns & 3,878 & 1,247  & 306  & 49,460 \\
\%Uniq SysTurns  & 76.6\%& 78.4\% & 2.6\% & 87.2\%\\
\end{tabular}}
\vspace{-6mm}
\end{table}
\vspace*{0pt}

%% file: experiments.tex

\textbf{Experimental Setup:} Our model architecture consists of four encoders: the utterance
encoder, hierarchical dialogue encoder, past DAs encoder and an agent
encoder. Our utterance encoder is a bi-directional LSTM with hidden layer size of
128. The utterance representation is the final state of the biLSTM. The
hierarchical dialogue encoder is an LSTM which takes the utterance representation
as input and its hidden size is 256. The past DAs vector is a
concatenation of the many-hot representations of past DAs wherein each
DA is many-hot over a set of 20 DAs. The agent encoding is
an indicator number representing the agent of the turn - 0 for the user, 1 for
the system.  We concatenate these representations and pass it through a
feed-forward network to make a binary decision per DA. For training,
we use ADAM for optimization with a learning rate of 0.001 and default
parameters. Our batch size is 100 for training. We initialize our word
embeddings with pretrained fastText ~\cite{bojanowski2017enriching} embeddings and fine-tune during training.

%% file: alignment.tex
\vspace{-2mm}
\section{Universal DA Schema}

\vspace*{0pt}
\subsection{Union of acts based on namespace}
\vspace{-2mm}
In order to align the respective acts in the datasets (GSim and DSTC2), we first took a
union of all the acts based on their names to create a unified representation. 
Figure ~\ref{fig:distribution} represents the
distribution of DAs used for the system side in these
datasets. Since our final aim is to tag human-human conversations (MultiWOZ-2.0 \cite{multiwoz}) with our 
unified set of acts, we have also included the distribution of acts in MultiWOZ-2.0 
for completeness, after stripping off the domain-name from the acts.
 It can be observed from the distribution that apart from a
few common acts like \textit{inform}, {\textit{request} etc., these datasets do not
share the same namespace for DAs, and even when the names
are the same, there may be differences in their semantics, as the
distributions of the acts are very different. For example, MultiWOZ-2.0
does not include the \textit{offer} act, whereas it appears in about 7\% of
the system turns for GSim and 50\% of the system turns for
DSTC2. Similarly, GSim and MultiWOZ-2.0 both have $select$, about 20\%
and 5\% of the turns respectively, whereas it is observed rarely in
DSTC2.
\vspace*{0pt}

\vspace*{0pt}
\begin{figure}[]
\vspace*{0pt}
  \caption{Distribution of system acts across datasets}  \label{fig:distribution}
\vspace{-4mm}
\hspace{-6mm}
  \begin{minipage}{0.40\textwidth}
    \includegraphics[width=1.3\textwidth]{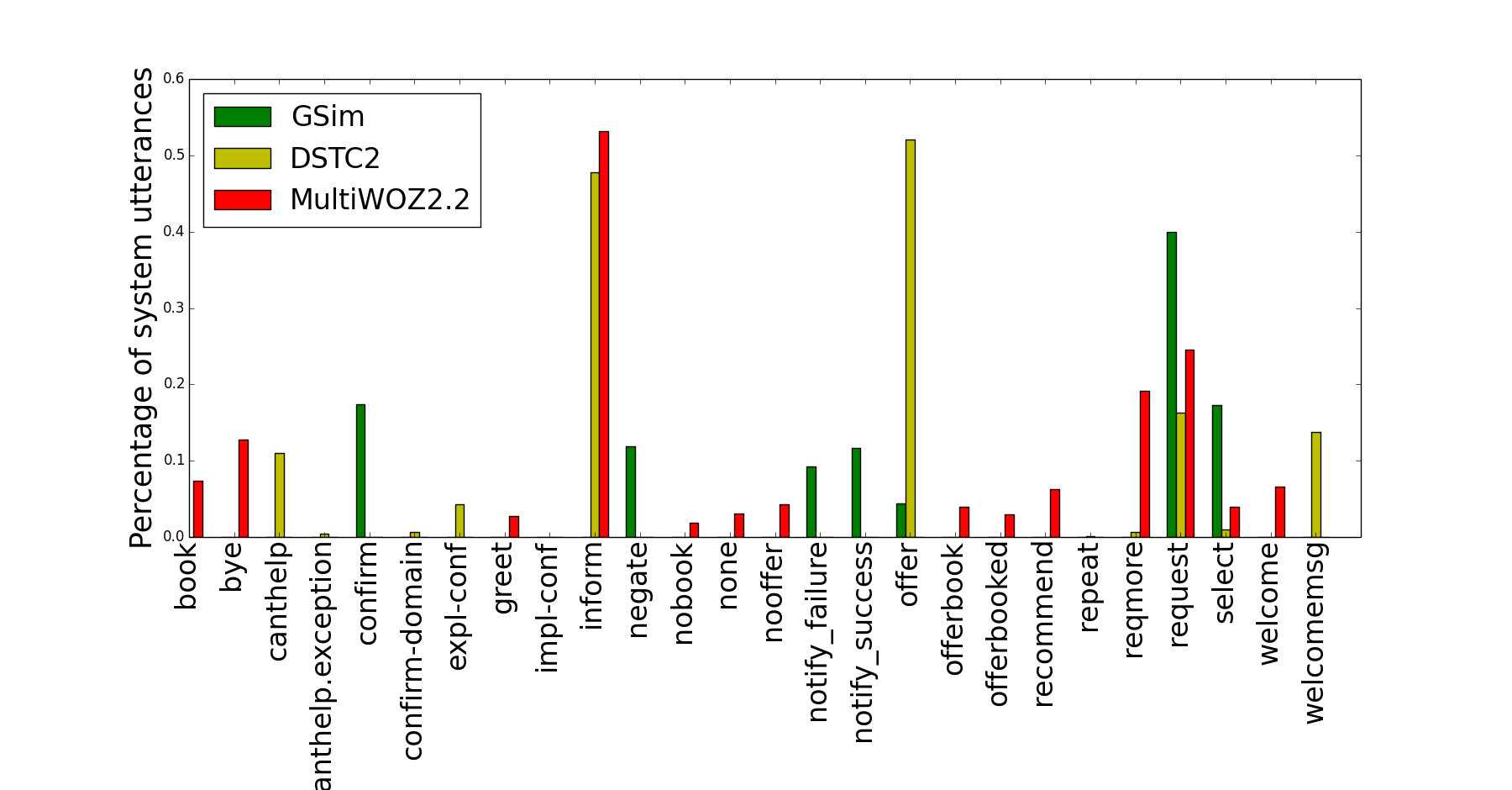}
  \end{minipage}\hfill
\end{figure}
\vspace*{0pt}

\vspace{-2mm}
\subsection{Tackling Annotation Mismatch: Manual alignments}
\vspace*{0pt}
Due to the lack of a shared namespace of acts, we manually assessed the semantics
of the acts in the datasets and found some obvious alignments. 
Table~\ref{table:manual_alignment_table} includes example alignments.

\begin{table}[]
\vspace{-1mm}
   \caption{Examples of manual alignment of acts in all datasets.}
   \label{table:manual_alignment_table}
   \vspace{-2mm}
   \resizebox{\columnwidth}{!}{
  \begin{tabular}{lll||l}
    GSim & DSTC2 & MultiWOZ-2.0 &Univ DA Schema\\\hline\hline
    notify\_failure & canthelp.exception & NoBook & sys\_notify\_failure\\\hline
    confirm & confirm-domain & & sys\_expl\_confirm\\
    & expl-conf & &\\\hline
    offer & offer & Recommend & offer\\\hline
  \end{tabular}}
  \vspace{-3mm}
\end{table}

Post-alignment, many acts in these datasets were shared between the user and
system such as \textit{inform} and \textit{negate}. However, we observed that these acts do not
share the same semantics and wording and hence the flow of the conversation varies based on which
agent the turn belongs to. Thus, to curate our unified schema of acts -
we made a finer distinction between user/system acts i.e. \textit{negate} from the
user is a \textit{user-negate} whereas from system, is \textit{system-negate}. Finally, we
train our DA tagger with the manually-aligned DSTC2 and GSim data.

To gauge the effectiveness of manual-alignments, we trained our
DA tagger on one dataset and tested it on the other to
see the inter-dataset and intra-dataset confusion. The results of
these experiments are listed in Table~\ref{table:baseline} as \textit{Baseline} numbers. The best
result for each test-set is highlighted in the table. As expected, for each test-set, we
obtain the best F1 scores when we use the matching training-set. On
the combined test-set, the model trained after combining all the datasets performs the best.


\vspace*{0pt}
\begin{table}
\vspace{-4mm}
\caption{Comparison of models trained on manually-aligned(baseline)
 vs the final machine-aligned universal schema (univ) based on F1 scores.
 Inter-dataset numbers represent the setting where the train/test data belong to
  different datasets. In intra-dataset, they belong to the same one. 
  `All' is the combination of respective partitions of all datasets}\label{table:baseline}.
\vspace{-3mm}

\resizebox{\columnwidth}{!}{
\begin{tabular} {|l|l|c|c|c||l|}
\hline
\multicolumn{2}{|c|}{} &\multicolumn{4}{c|}{Training Set}\\\hline
\multicolumn{2}{|c|}{Test Sets}  &Gsim-R & Gsim-M & DSTC2 & All \\
\hline
\multirow{2}{*}{Gsim-R} & Baseline & {\bf 0.867} & 0.706 & 0.324 & {\bf 0.897}\\
                        &Univ  & {\bf 0.892} & 0.751 & 0.453& {\bf 0.916} (U-DAT)\\
\multirow{2}{*}{Gsim-M}&Baseline & 0.801 & {\bf 0.904} & 0.382 & {\bf 0.908}\\
                      & Univ  & 0.850 & {\bf 0.914} & 0.474& {\bf 0.921} (U-DAT)\\

\multirow{2}{*}{DSTC2}&Baseline & 0.434 & 0.365 & {\bf 0.909} & 0.899\\
                       & Univ  & 0.564 & 0.496 & {\bf 0.920}& 0.917 (U-DAT)\\\hline\hline
\multirow{2}{*}{All}& Baseline & 0.560 & 0.477 & 0.742 & {\bf 0.900}\\
                      & Univ & 0.659 & 0.583 & 0.786 & {\bf 0.921} (U-DAT)\\
\hline
\multicolumn{4}{|c|}{Avg of inter-dataset scores (Baseline/Univ)}  & \multicolumn{2}{|c|}{0.439/0.555}\\
\multicolumn{4}{|c|}{Avg of intra-dataset scores (Baseline/Univ)} & \multicolumn{2}{|c|}{0.898/0.912}\\\hline

\end{tabular}}
\vspace{-2mm}
\end{table}

\subsection{ Machine-aided alignments}
\vspace*{0pt}
After manually aligning the acts across datasets, we still observed
poor performance on the task.  Looking at the various training and
validation set DAs in the manually curated unified
representation, we noticed some semantically similar acts which were
confusing our tagger. Some examples are:
\vspace*{0pt}
\begin{enumerate}[topsep=0pt,itemsep=-1ex,partopsep=1ex,parsep=1ex]
\vspace*{0pt}
\item \textbf{Mod1: offer/select}- {\em I found a show for 7.30 pm}/{\em I found shows for
  5 pm and 7 pm}. We merge these acts.

\item \textbf{Mod2: user-request/sys-request}- {\em What is the phone number?/What kind of food would you
  like?} We merge these acts.

\item \textbf{Mod3: affirm(x=y)/affirm + inform(x=y)}- \textit{affirm} with
  slots is equivalent to separate \textit{affirm} and \textit{inform} DAs, for eg. `yes, 7pm'
  can become \textit{affirm, inform(time=7pm)} from \textit{affirm(time=7pm)}. We split them.

\item \textbf{Mod4: reqalts/reqmore}- \textit{Is there anything else?/Can i help you with
  anything else?} We merge these acts.

\end{enumerate}
We merged/split DAs like the aforementioned ones, as they
can easily be restored using other information. For example, if
multiple results are offered, we could convert an \textit{offer} act
to a \textit{select} act, or depending on the agent, we can
convert a \textit{request} act to a \textit{user-request} or a
\textit{sys-request}.  The effect of these transformations on inter and intra-dataset F1 scores is shown in
Table~\ref{table:alignment}. 

\begin{table}[]
   \caption{Effect of squashing/splitting different acts on inter and 
   intra-dataset average F1 score. Each column displays
     the effect of addition of modification to the one on its left.}\label{table:alignment}
   \vspace*{0pt}
   \resizebox{\columnwidth}{!}{
  \begin{tabular}{|l|c|c|c|c|c|}\hline
    & manually-aligned & +Mod1 & +Mod2 & +Mod3 & +Mod4\\\hline

    Inter-dataset  &0.439 &0.436 &0.512 &0.527 &0.555\\

    Intra-dataset  & 0.898 &0.903 &0.896 &0.890 &0.912\\ \hline
  \end{tabular}}
  \vspace{-4mm}
\end{table}

After performing all these transformations\footnote{Details in Appendix, Table~\ref{tab:data_das}}, we curated a Universal DA schema of
20 acts which capture the entirety of all the acts present in these datasets. We
present these in Table~\ref{table:list}. We compare the F1 scores of DA tagging
models trained using this schema with our original baseline (models trained on
manually-aligned acts) in Table~\ref{table:baseline}.  The best-performing model
is obtained by combining all the GSim and DSTC2 datasets using the Universal DA
schema. We refer to this model as U-DAT.

\begin{table}[]
\vspace{-5mm}
  \caption{Universal DA schema}\label{table:list}
\vspace{-3mm}
\vspace*{0pt}

\begin{tabular}{|p{7.5cm}|}
  \hline
  \textit{ack, affirm, bye, deny, inform, repeat, reqalts, request, restart, thank-you, user-confirm, sys-impl-confirm, sys-expl-confirm, sys-hi, user-hi, sys-negate, user-negate, sys-notify-failure, sys-notify-success, sys-offer}\\\hline
\end{tabular}
\vspace{-4mm}
\end{table}

%% file: results.tex
\vspace{-2mm}
\section {DA Tagging of Human-Human Datasets}
For experimenting with DA annotation of human-human (HH) dialogues, we
used MultiWOZ-2.0\cite{multiwoz} as our dataset. This version of the dataset only has DAs for the
system turns. 

To do an evaluation on MultiWOZ-2.0, 
we first need to map the dataset to our Universal DA schema. The distribution
 of acts in MultiWOZ-2.0 can be seen in Figure~\ref{fig:distribution}.
However, during manual assessment, we found that while most of the acts in MultiWOZ-2.0
 dataset aligned well with our Universal DA schema, the
annotations in \textit{inform/select/recommend} and
\textit{general-domain} space of acts were inconsistent in
MultiWOZ-2.0. 
For example, \textit{select} was often confused with \textit{inform}.
Additionally, the MultiWOZ-2.0 act annotation space lacks granularity
for expressing intent. Thus, to maximally align MultiWOZ-2.0 with the Universal DA schema, we use heuristics. For
example, we check for presence of keywords like `bye', `thank you' etc. to
label the \textit{bye} and \textit{thank\_you} class of system
acts\footnote{Details in Appendix, Table~\ref{tab:heuristics}}. To evaluate the
effectiveness of our heuristics, we manually annotated a smaller subset
(524 turns) of the MultiWOZ-2.0 test-set with DAs in our Universal DA schema,
we call this as the \textit{univ-testset}. We then trained 2 DA tagging
models on MultiWOZ-2.0 - one with the DA labels mapped to the
 Universal DA schema using heuristics (say \textit{heuristics-model}) and one
without (say \textit{no-heuristics-model}). On \textit{univ-testset}, we got 
an F1 score of 0.609 with \textit{no-heuristics-model} and 0.716 with 
\textit{heuristics-model}, which validates the effectiveness of our heuristics.

Due to labeling inconsistency in MultiWOZ-2.0 as described above, to do an
accurate evaluation on HH datasets, we use \textit{univ-testset} as our test-set
henceforth. For training and validation, we use the standard dataset partitions.

\vspace{-2mm}
\subsection{Adaptation to Human-Human dialogues}
\vspace{-1mm}
In addition to the U-DAT model, which is unsupervised with respect to HH data, we train
2 other models\footnote{Due to the absence of user-acts in MultiWOZ-2.0, we removed the past 
DA encoder from the context encoder of both the models.} in semi-supervised and supervised settings.
\vspace{-1mm}
\begin{itemize}[labelindent=0em,leftmargin=1em]
\itemsep -0.5ex
\item {\bf Semi-supervised HH U-DAT:} Our aim is to see the quality of DA
  annotations without any labeled HH dialogues. For this, we
  labeled the MultiWOZ-2.0 corpus with U-DAT. Then, we trained another
  model with the estimated DA labels. This model is semi-supervised,
  as it doesn't use any manually labeled data, but uses the labels generated by U-DAT.
\item {\bf Supervised HH U-DAT:} We trained a supervised DA tagging model with the manually
  annotated DA labels mapped to our Universal DA schema.
\end{itemize}
\vspace{-1mm}

\begin{figure}
\vspace{-2mm}
          \caption{DA tagging: System-side F1 score learning curves
            using U-DAT and models trained using HH data on \textit{univ-testset}.}\label{fig:learning_curve}
     \vspace{-2mm}
     \hspace{-5mm}
    \begin{minipage}{0.40\textwidth}
    \vspace{-2mm}
        \includegraphics[width=1.3\textwidth]{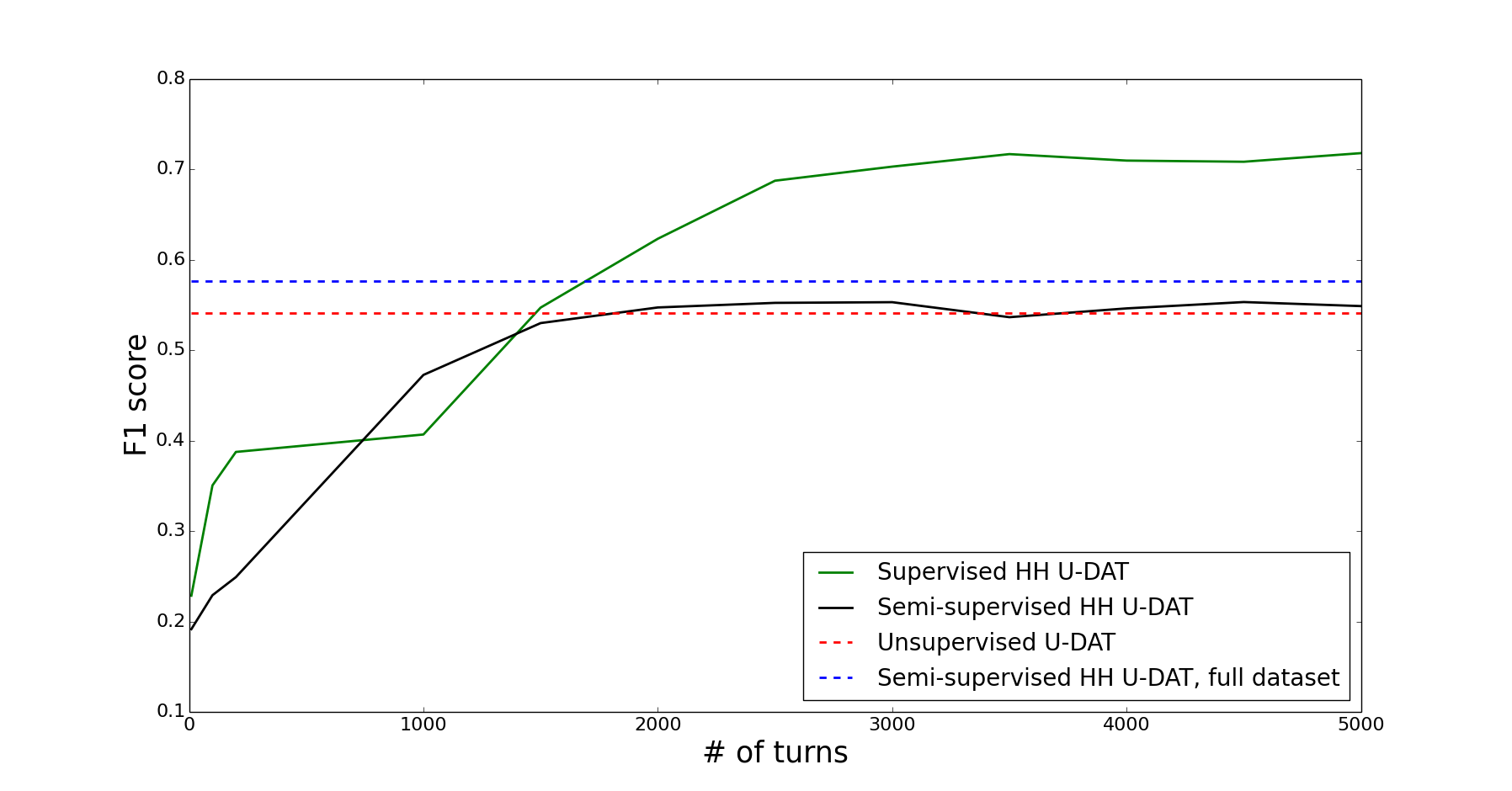} %
     \vspace{-2mm}
    \end{minipage}\hfill
\vspace{-2mm}
\end{figure}

We plot learning curves by varying the amount of data used in each
model in Figure~\ref{fig:learning_curve}. The black curve shows the effect of varying the amount of data on the performance of {\bf Semi-supervised HH U-DAT}. The blue line is the performance of {\bf Semi-supervised HH U-DAT} 
with \textit{all} the MultiWOZ-2.0 training dialogues, a system-side F1 score of
0.577. The red line
corresponds to the performance of {\bf U-DAT}, a system-side F1 score
of 0.541. The red and blue lines show that, if unannotated data is available
from the HH conversations, we can improve the DA
tagging F1 score by 3.6\% absolute. As can be seen from the green
curve obtained with {\bf Supervised HH U-DAT}, we would need over 1700
manually annotated examples to reach the best semi-supervised learning
F1 score. This provides useful guidelines on the amount of data required
for accurate DA tagging.

\vspace{-2mm}
\subsection{Analysis of domain adaptation via self-training}
\vspace{-1mm}
To gauge the extent of domain adaptation by self-training (semi-supervised) over HH data,
 we performed leave-one-domain-out experiments. For each domain X, we 
 train 3 models on HH data (mapped to Universal DA schema) with the following settings:
\vspace*{0pt}
\begin{itemize}[labelindent=0em,leftmargin=1.2em]
\vspace*{0pt}
 	\itemsep -0.5ex
 	\item \textit{HH-UDAT, no domain data, supervised}: We use 3000 
 	turns of manually-labeled out-of-domain(OOD) data i.e. we exclude turns from domain X.
 	Since the data is manually-labeled, this model is supervised.
 	\item \textit{HH-UDAT, w-domain data, semi-supervised}: In addition to OOD data used above, 
 	we use 300 turns of data from X labeled using U-DAT. This model is 
 	semi-supervised w.r.t X.
 	\item \textit{HH-UDAT, w-domain data, supervised}: In addition to OOD data, 
 	we use the manual-labels of the 300 turns of X data used above.
 	This model is supervised w.r.t X.
 \end{itemize}
 \vspace{-1mm}

\begin{table}[]
\vspace{-2mm}
   \caption{Per-domain F1 score results for domain adaptation via self-training (semi-supervised training)}\label{table:domain_ablation}
   \vspace{-2mm}
   \scalebox{0.9}{
    \begin{tabular}{|l|c|c|c|c|}\hline
    Domain 	& U-DAT & 	HH-UDAT 	& HH-UDAT  & HH-UDAT\\
    Domain data	& N/A &	no & yes  	&  yes\\ 
    Supervision	& no & yes & semi & yes\\\hline
    Restaurant & 0.613 & 0.703 & 0.721 & 0.735\\ 
    Hotel &  0.571 & 0.685 & 0.697 & 0.701 \\
    Train & 0.523 & 0.723 & 0.666 &  0.724\\
    Taxi & 0.709 & 0.727 & 0.787 & 0.784 \\
    Attraction & 0.444 & 0.672 & 0.689 & 0.728 \\\hline
    Average & 0.572 & 0.702 & 0.712 & 0.734\\\hline
  \end{tabular}}
  \vspace{-6mm}
\end{table}

The results of these experiments are listed in
Table~\ref{table:domain_ablation}.  The results show improvements when
unlabeled (0.712 vs 0.702) or labeled (0.734 vs 0.702) target domain data is
available.
\vspace{-2mm}

%% file: conclusion.tex
\section{Conclusions}
\vspace{-1mm}
\label{sec:conclusion}
We are interested in DA tagging of human-human conversations
with the final goal of end-to-end training of task-oriented dialogue
systems, so that we can generate system actions for a given dialogue
context.  In this work, we investigated multiple annotated
human-machine conversation datasets, with differences in DA
schema.  We discussed manual and automatic approaches for aligning
these different schemas, and presented results on a target corpus of
human-human dialogues. We demonstrated that without manually
annotating any new human-human conversations, we achieve an F1 score of 57.7\%, which requires at least 1.7K turns of manually annotated
human-human dialogue data. We provided learning curves to present
performance improvement with different amounts of manually and
automatically labeled data which provides useful guidelines 
on the amount of data required for accurate DA tagging. In the presence of a new domain, we compared the performance of DA tagging using unsupervised,
semi-supervised and supervised approaches. For these domains, we
showed further improvements when unlabeled or labeled target domain data is
available. As future work, we intend to further explore domain adaptation and use these annotated human-human conversations to train end-to-end task-oriented dialogue systems.

%% file: appendix.tex
\begin{appendices}
\section {Appendix}

\begin{table*}[h]
\caption{Alignment of Datasets with Universal DA Schema}
\label{tab:data_das}

\resizebox{\textwidth}{!}{
  \begin{tabular}{|c|c|c||c|}
  \hline
GSim-R & GSim-M & DSTC2 & Universal DA Schema\\
\hline
\hline
  inform(x=y) & inform(x=y) & inform(x=y) & inform(x=y)\\
 request(x) & request(x) & request(x) & request(x)\\
   negate(x=y) & negate(x=y) & negate(x=y) & user-negate(x=y)\\
   negate(x=y) & negate(x=y) & negate(x=y) & sys-negate(x=y)\\
   & & hello() & user-hi()\\
   greeting(x=y) & greeting(x=y) & & user-hi() + inform(x=y)\\
    &  & welcomemsg() & sys-hi()\\
   request\_alts() & request\_alts() & reqalts() & reqalts()\\
   request\_alts() + inform(x=y) & request\_alts() + inform(x=y) & reqalts() + inform(x=y) & reqalts(x=y)\\
  & & reqmore() & reqalts()\\
   cant\_understand() & cant\_understand() & & repeat()\\
    &  & repeat() & repeat()\\
    &  & restart() & restart()\\
   affirm() & affirm() & affirm() & affirm()\\
   affirm(x=y) & affirm(x=y) & & affirm() + inform(x=y)\\
   & & impl-conf(x=y) & sys-impl-confirm(x=y)\\
   confirm(x=y) & confirm(x=y) & expl-conf(x=y), confirm-domain(x=y) & sys-expl-confirm(x=y)\\
   & & confirm(x=y) & user-confirm(x=y)\\
   notify\_failure() & notify\_failure() & canthelp(), canthelp.exception() & sys-notify-failure()\\
   notify\_success() & notify\_success() &  & sys-notify-success()\\
   offer(x=y) & offer(x=y) & offer(x=y) & sys-offer(x=y)\\
   select(x=y1,y2) & select(x=y1,y2) & select(x=y1,y2) & sys-offer(x=y1,y2)\\
   thank\_you() & thank\_you() & thankyou() & thank\_you()\\
   & & bye() & bye()\\
   & & ack() & ack()\\
   & & deny(x=y) & deny(x=y)\\
   \hline

\end{tabular}}

\end{table*}

\begin{table*}[h]
\caption{Heuristics to fix MultiWOZ2.0 and align it with the Universal DA Schema}
\label{tab:heuristics}

\resizebox{\textwidth}{!}{
  \begin{tabular}{|c|c||c|}
  \hline
    MultiWOZ-2.0 acts & Heuristic & Universal DA Schema\\
    \hline
    \hline
    'Booking-Request','Restaurant-Request','Hotel-Request' & & request\\
    'Attraction-Request','Taxi-Request','Train-Request' & & \\\hline
    'Train-OfferBook', 'Booking-Inform' & & sys-expl-confirm\\\hline
    'Booking-NoBook','Restaurant-NoOffer', 'Hotel-NoOffer' & & sys-notify-failure\\
    'Attraction-NoOffer','Train-NoOffer' & & \\\hline
    'Booking-Book', 'Train-OfferBooked & & sys-notify-success\\\hline
    'Restaurant-Recommend', 'Restaurant-Select', 'Hotel-Recommend', 'Hotel-Select' & &sys-offer\\
    'Attraction-Recommend', 'Attraction-Select', 'Train-Select' & &\\\hline
    'general-reqmore' & & reqalts\\\hline
    'general-greet' & contains '*bye' & bye \\
    'general-bye' & & \\
    'general-welcome' & & \\\hline
    'general-greet' & contains '*thank*' & thank\_you \\
    'general-welcome' & & \\\hline
    'Restaurant-Inform','Hotel-Inform' & ends in 'Booking-Inform(none=none)' & sys-offer\\
    'Attraction-Inform','Taxi-Inform','Train-Inform' &  & \\\hline
    'Restaurant-Inform','Hotel-Inform' & doesn't end in 'Booking-Inform(none=none)' & inform\\
    'Attraction-Inform','Taxi-Inform','Train-Inform' & & \\\hline
\end{tabular}}

\end{table*}

\end{appendices}